\documentclass[11pt,a4paper]{article}
\usepackage{emnlp-ijcnlp-2019}
\usepackage{times}
\usepackage{latexsym}
\usepackage{todonotes}
\usepackage{tikz-dependency}
\usepackage{subfig}
\usepackage{pgfplots}
\usepackage{xcolor}
\usepackage{pgfkeys}
    \newenvironment{customlegend}[1][]{%
        \begingroup
        \csname pgfplots@init@cleared@structures\endcsname
        \pgfplotsset{#1}%
    }{%
        \csname pgfplots@createlegend\endcsname
        \endgroup
    }%
    \def\addlegendimage{\csname pgfplots@addlegendimage\endcsname}

\usepackage{url}
\usepackage{algorithm2e}
\usepackage{booktabs}
\usepackage{multirow}
\usepackage{graphicx}
\usepackage{amsmath}
\usepackage{algorithm2e}
\aclfinalcopy % Uncomment this line for the final submission

\def\bb{\mathbf{b}}
\def\bc{\mathbf{c}}

\def\be{\mathbf{e}}

\def\bg{\mathbf{g}}

\def\bp{\mathbf{p}}

\def\bs{\mathbf{s}}

\def\bu{\mathbf{u}}
\def\bv{\mathbf{v}}

\def\bx{\mathbf{x}}

\def\bW{\mathbf{W}}

\def\cL{\mathcal{L}}

\def\cR{\mathcal{R}}

\def\cT{\mathcal{T}}

\DeclareMathOperator*{\softmax}{softmax}

\DeclareMathOperator*{\BiLSTM}{Bi-LSTMs}

\DeclareMathOperator*{\Lookup}{Lookup}
\DeclareMathOperator*{\Squash}{Squash}
\DeclareMathOperator*{\sign}{sign}

\title{Capturing Argument Interaction in \\ Semantic Role Labeling with Capsule Networks}

 \author{{Xinchi Chen}$^1$ ~~ {Chunchuan Lyu}$^1$ ~~ {Ivan Titov}$^{1,2}$ \\
 xchen13@exseed.ed.ac.uk ~~ chunchuan.lv@gmail.com ~~ ititov@inf.ed.ac.uk  \\
        $^1$ {ILCC, School of Informatics, University of Edinburgh}  \\
$^2${ILLC, University of Amsterdam} 
}

\date{}

\begin{document}
\maketitle

\begin{abstract}

Semantic role labeling (SRL) involves extracting propositions (i.e. predicates and their typed arguments) from natural language sentences. State-of-the-art SRL models rely on powerful encoders (e.g., LSTMs) and do not model non-local interaction between arguments. We propose a new approach to modeling these interactions while maintaining efficient inference. Specifically, we use  Capsule Networks \cite{hinton2017capsulenet}: each proposition is encoded as a tuple of  \textit{capsules}, one capsule per argument type (i.e. role). These tuples serve as embeddings of entire propositions.   In every network layer, the capsules interact with each other and with representations of words in the sentence.  Each iteration results in updated proposition embeddings and updated predictions about the SRL structure. Our model substantially outperforms the non-refinement baseline model on all 7 CoNLL-2019 languages and achieves state-of-the-art results on 5 languages (including English) for dependency SRL. We analyze the types of mistakes corrected by the refinement procedure. For example,  each role is typically (but not always) filled with at most one argument. Whereas enforcing this approximate constraint is not useful with the modern SRL system, iterative procedure corrects the mistakes by capturing this intuition in a flexible and context-sensitive way.\footnote{Code:  https://github.com/DalstonChen/CapNetSRL.}

    %Previous state-of-the-art dependency semantic role labeling models have limited interactions between arguments, leading to independent argument prediction. In this paper, we employ capsule network to encode the interactions between arguments for each predicate. We represent each argument role candidate as a capsule to encode more information. Capsules interact and collaborate from each other via capsule network layer. To better incorporate more interaction globally, we further introduce a global node to capture more high level signals and each capsule exchanges information with global node. Experiment results show the effectiveness of our model.
\end{abstract}

\section{Introduction}

The task of semantic role labeling (SRL) involves the prediction of predicate-argument structure, i.e., both the identification of arguments and labeling them with underlying semantic roles.  
The shallow semantic structures have been shown beneficial in many natural language processing (NLP) applications, including information extraction~\cite{Christensen:2011:AOI:1999676.1999697}, question answering~\cite{eckert-neves-2018-semantic} and machine translation~\cite{marcheggiani-etal-2018-exploiting}. 

\begin{figure}[t]
    \centering
    \includegraphics[width=0.48\textwidth]{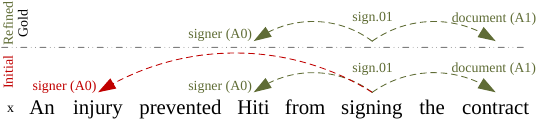}
    \caption{An example predicate-argument structure: green and red arcs denote correct and wrong predictions, respectively.}
    \label{fig:example}
\end{figure}

In this work, we focus on the dependency version of the SRL task~\cite{Surdeanu:2008:CST:1596324.1596352}. An example of a dependency semantic-role structure is shown in Figure~\ref{fig:example}. Edges in the graph are marked with semantic roles, whereas predicates (typically verbs or nominalized verbs) are annotated with their senses.  

Intuitively, there are many restrictions on potential predicate-argument structures. Consider the `role uniqueness' constraint: each role is typically, but not always, realized at most once. For example, predicates have at most one {\it agent}. Similarly, depending on a verb class, only certain subcategorization patterns are licensed.
Nevertheless, rather than modeling the interaction between argument labeling decisions,
state-of-the-art semantic role labeling models~\cite{marcheggiani2017encoding,cai2018full,li2019dependency} rely on powerful sentence encoders (e.g., multi-layer Bi-LSTMs~\cite{zhou-xu-2015-end,qian-etal-2017-syntax,Tan2018DeepSR}).  This contrasts with earlier work on SRL, which hinged on global declarative constraints on the labeling decisions~\cite{fitzgerald2015semantic,tackstrom-etal-2015-efficient,das2012exact}.  Modern SRL systems are much more accurate and  hence enforcing such hard and often approximate constraints is unlikely to be as beneficial (see our experiments in Section \ref{sec:model_select}). 

Instead of using hard constraints, we propose to use a simple iterative structure-refinement procedure.
It starts with independent predictions and refines them in every subsequent iteration. When refining a role prediction for a given candidate argument, information about the assignment of roles to other arguments gets integrated. Our intuition is that modeling interactions through the output spaces rather than hoping that the encoder somehow learns to capture them, provides a useful inductive bias to the model. In other words, capturing this interaction explicitly should mitigate overfitting.
This may be especially useful in a lower-resource setting but, as we will see in experiments, it appears beneficial even in a high-resource setting.

% I use a hypothetical example
% An injury prevented Hiti from signing the contact
% Maybe find a better example where there is a potential for a mistake

We think of semantic role labeling as extracting propositions, i.e. predicates and argument-role tuples. In our example, the proposition is \texttt{sign.01}(\texttt{Arg0:signer} = \textit{Hiti}, \texttt{Arg1:document} = \textit{contract}). 
Across the iterations, we maintain and refine not only predictions about propositions but also
their  embeddings. Each `slot' (e.g., correspond to role \texttt{Arg0:signer}) is represented with a vector, encoding information about arguments assumed to be filling this slot. For example, if  \textit{Hiti} is predicted to fill slot \texttt{Arg0}  in the first iteration then the slot representation will be computed based on the contextualized representation of that word and hence reflect this information. The combination of all slot embeddings (one per semantic role) constitutes the embedding of the proposition. The proposition embedding, along with the prediction about the semantic role structure, gets refined in every iteration of the algorithm. 

Note that, in practice, the predictions in every iteration are soft, and hence proposition embeddings will encode current beliefs about the predicate-argument structure.
The distributed representation of propositions provides an alternative (``dense-embedding'') view on the currently considered semantic structure, i.e. information extracted from the sentence. Intuitively, this representation can be readily tested to see how well current predictions satisfy selection restrictions (e.g., \textit{contract} is a very natural filler for {\tt Arg1:document}) and check if the arguments are compatible. 

To get an intuition how the refinement mechanism may work,
imagine that both \textit{Hiti} and \textit{injury} are labeled as {\tt Arg0}  for {\tt sign.01} in the first iteration. Hence, the representation of the slot {\tt Arg0} will encode information about both predicted arguments.
At the second iteration, the word \textit{injury} will be aware that there is a much better candidate for filling the {\tt signer} role, and the probability of assigning \textit{injury}  to this role will drop.
As we will see in our experiments, whereas enforcing the hard uniqueness constraint is not useful,  our iterative procedure corrects the mistakes of the base model by capturing interrelations between arguments in a flexible and context-sensitive way.

In order to operationalize the above idea, we take inspiration from the Capsule Networks~(CNs)~\cite{hinton2017capsulenet}. Note that we are not simply replacing Bi-LSTMs with generic CNs. Instead, we use CN to encode the structure of the refinement procedure sketched above. Each slot embedding is now a \textit{capsule} and each  proposition embedding is  a tuple of  capsules, one capsule per role. In every iteration (i.e. CN network layer), the capsules interact with each other and with representations of words in the sentence. 
%Besides using the slot-specific capsules, we use capsules to represent refined representations of words and experimented with an extra capsuled encoding the global information about the proposition.

% We experiment with our model on standard benchmarks for two languages, English and Japanese. The two datasets differ greatly in their size, with the English dataset being about ten times larger. Our model achieves the best-reported results in both cases, and we observe substantial improvements from using the iterative procedure. 
We experiment with our model on standard benchmarks for 7 languages from CoNLL-2019. Compared with the non-refinement baseline model, we observe substantial improvements from using the iterative procedure. The model achieves state-of-the-art performance in 5 languages, including  English.

\section{Base Dependency SRL Model}\label{sec:base}
In dependency SRL, for each predicate $p$ of a given sentence $x = \{x_1, x_2, \cdots, x_n\}$ with $n$ words, the model needs to predict roles $y = \{y_1, y_2, \cdots, y_n\}$ for every word. The role can be {\it none}, signifying that the corresponding word is not an argument of the predicate.
%The aim of SRL is to maximize the probability of role predictions $P(y)$:

We start with describing the factorized baseline model which is similar to that of \citet{marcheggiani2017simple}. %, which we use as the first step in refinement \xchen{Is it the first step of our refinement process?} and also as the baseline.
 %, though, as we will see in our experiments, more accurate.
It consists of three components: (1) an embedding layer; (2) an encoding layer and (3) an inference layer.

\subsection{Embedding Layer}
The first step is to map symbolic sentence $x$ and predicate $p$ into continuous embedded space:% by using static embeddings or expense \cite{peters-etal-2018-deep} (see Section~\ref{?????}): \Ivan{What do YOU do? This should not be a hypothetical system but our base model}
\begin{align}
    \be_i &= \Lookup (x_i), \\
    \bp &= \Lookup (p), 
\end{align}
where $\be_i \in \cR^{d_e}$ and $\bp \in \cR^{d_e}$.
\subsection{Encoding Layer}
The encoding layer extracts features from input sentence $x$ and predicate $p$. We extract features from input sentence $x$ using stacked bidirectional LSTMs \cite{Hochreiter:1997:LSM:1246443.1246450}:
\begin{equation}
    \bx_i = \BiLSTM (\be_i),
\end{equation}
where $\bx_i \in \cR^{d_l}$.  Then we represent each role logit of each word by a bi-linear operation with the target predicate:
\begin{equation}
    b_{j|i} = \bx_i  \bW  \bp, \label{eq:logit}
\end{equation}
where $\bW \in \cR^{d_l \times d_e}$ is a trainable parameter and $b_{j|i} \in \cR$ is a scalar representing the logit of role $j$ for word $x_i$.
\subsection{Inference Layer}
The probability $P(y|x, p)$ is then computed as
\begin{align}
    P(y|x, p) &= \prod_i P(y_i|x, i, p), \label{eq:prob}\\
    & = \prod_i \softmax(\bb_{\cdot|i})_{y_i},
\end{align}
where $\bb_{\cdot|i} = \{b_{1|i}, b_{2|i}, \cdots, b_{|\cT||i}\}$ and $|\cT|$ denotes the number of role types.
\section{Dependency based Semantic Role Labeling using Capsule Networks}\label{sec:capsule}
\begin{figure}[t]
    \centering
    \includegraphics[width=0.48\textwidth]{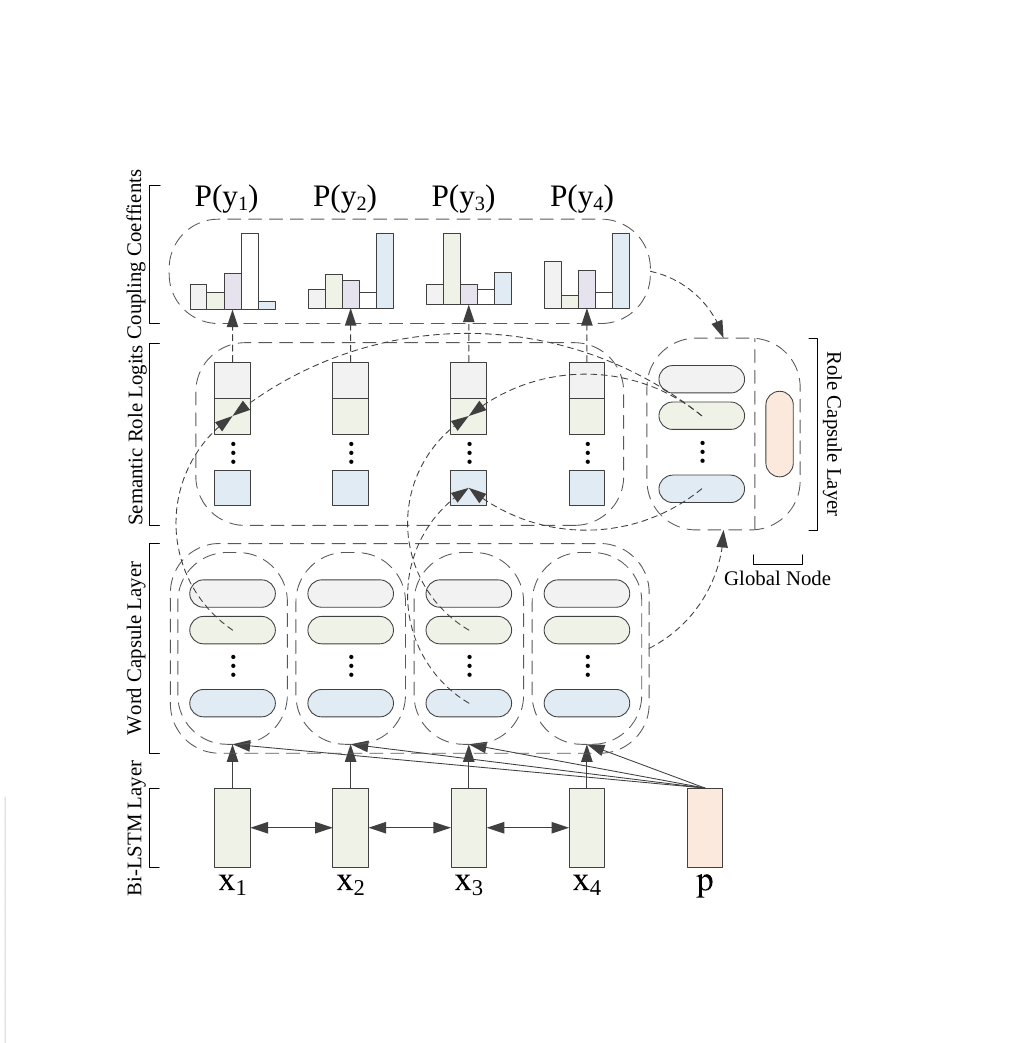}
    \caption{Architecture of the proposed CapsuleNet with the global capsule node. The dashed arcs are iterable.}
    \label{fig:capsule}
\end{figure}

Inspired by capsule networks, we use capsule structure for each role state to maintain information across iterations and employ the dynamic routing mechanism to derive the role logits $b_{j|i}$ iteratively. Figure \ref{fig:capsule} illustrates the architecture of the proposed model.

\subsection{Capsule Structure}
%In order to derive the role logits $b_{j|i}$,
We start by  introducing two capsule layers: (1) the word capsule layer and (2) the role capsule layer. 
%As discussed in the introduction,  the union of roles capsules can be regarded as  representing an embedding of the proposition.

\subsubsection{Word Capsule Layer}

The word capsule layer is comprised of capsules representing the roles of each word. Given sentence representation $\bx_i$ and predicate embedding $\bp$, the word capsule layer is derived as:
\begin{equation}
    u_{j|i}^k = \bx_i  \bW_j^k  \bp; \,\,\,k = 1,2, \cdots, K, \label{eq:word}
\end{equation}
where $\bW_j^k \in \cR^{d_l \times d_e}$ and $\bu_{j|i} \in \cR^{K}$ is the capsule vector for role $j$ of word $x_i$. $K$ denotes the capsule size. Intuitively, the capsule encodes the argument-specific information relevant to deciding if role $j$ is suitable for the word. These capsules do not get iteratively updated.

\subsubsection{Role Capsule Layer}
As discussed in the introduction, the role capsule layer could be viewed as an embedding of a proposition. The capsule network generates the capsules in the  layer using ``routing-by-agreement''. This process can be regarded as a pooling operation. % and discards the location information, leading to a more robust network.
Capsules in the role capsule layer at $t$-th iteration are derived as:
\begin{align}
    \bv_j^{(t)} = \Squash (\bs_j^{(t)}) = \frac{||\bs_j^{(t)}||^2}{1 + ||\bs_j^{(t)}||^2} \frac{\bs_j^{(t)}}{||\bs_j^{(t)}||},
\end{align}
where intermediate capsule $s_j^{(t)}$ is generated with the linear combination of capsules in the word capsule layer with weights $c_{ij}^{(t)}$:
\begin{equation}
    \bs_j^{(t)}= \sum_i c_{ij}^{(t)}\bu_{j|i}.\label{eq:intermediate}
\end{equation}
Here, $c_{ij}^{(t)}$ are coupling coefficients, calculated by ``softmax'' function over role logits $b_{j|i}^{(t)}$ . The coefficients  $c_{ij}^{(t)}$ can be interpreted as  the probability that word $x_i$ is assigned role $j$:
\begin{equation}
    \bc_i^{(t)} = \softmax (\bb_{\cdot|i}^{(t)}). \label{eq:couple}
\end{equation}
The role logits $b_{j|i}^{(t)}$ are decided by the iterative \textit{dynamic routing} process. The $\Squash$ operation will deactive
capsules receiving small input ${\bs}_j^{(t)}$ (i.e. roles not predicted in the sentence) by pushing them further to the $\mathbf{0}$ vector.

\subsection{Dynamic Routing}
The dynamic routing process involves $T$ iterations. The role logits $b_{j|i}$ before first iteration are all set to zeros: $b_{j|i}^{(0)} = 0$. Then, the dynamic routing process updates the role logits $b_{j|i}$ by modeling agreement between capsules in two layers:
\begin{equation}
    b_{j|i}^{(t+1)} = b_{j|i}^{(t)} + \bv_j^{(t)} \bW \bu_{j|i}, \label{eq:dynamic}
\end{equation}
where $\bW \in \cR^{K \times K}$. %Here, $b_{j|i}^{(t)}$ denotes the role logits in $t$-th iteration. 

The whole dynamic routing process is shown in Algorithm \ref{alg:dynamic}. The dynamic routing process can be regarded as the role refinement procedure (see Section \ref{sec:refinement} for details).

\begin{algorithm}[t] \small
\SetAlgoLined
%\KwInput{\# of iterations $T$, all capsules in $l_w$ $\bu_{j|i}$}
%\KwResult{ semantic role logits $b_{j|i}^{(T)}$ }
\textbf{Input:} \# of iterations $T$, all capsules in $l_w$ $\bu_{j|i}$ \\
\textbf{Result:} semantic role logits $b_{j|i}^{(T)}$ \\
Initialization:  $b_{j|i}^{(0)} = 0$  \;
\For{$t \in [0, 1, \cdots, T-1]$}{
for all capsules in $l_w$: $\bc_i^{(t)} = \softmax (\bb_{\cdot|i}^{(t)})$\;
for all capsules in $l_r$: $\bs_j^{(t)}= \sum_i c_{ij}^{(t)}\bu_{j|i}$\;
for all capsules in $l_r$: $\bv_j^{(t)} = \Squash (\bs_j^{(t)})$\;
for all capsules in $l_w$ and $l_r$: $b_{j|i}^{\!(t+\!1)}\!\!\! = \!\! b_{\!j|i}^{(t)}\!\! + \!\bv_j^{\!(t)} \!\bW\! \bu_{\!j|i}$\;
}
\caption{Dynamic routing algorithm. $l_w$ and $l_r$ denote word capsule layer and role capsule layer, respectively.}
\label{alg:dynamic}
\end{algorithm}

\section{Incorporating Global Information}
When computing the $j$-th role capsule representation (Eq~\ref{eq:intermediate}), the information
originating from an $i$-th word (i.e. $u_{j|i}$) is weighted by the probability of assigning role $j$ to word $x_i$ (i.e. $c_{ij}$).
In other words, the role capsule receives messages only from words assigned to its role.  
 This implies that the only interaction the capsule network can model is
 competition between arguments for a role.\footnote{In principle, it can model the opposite, i.e. collaboration / attraction but it is unlikely to be useful in SRL.}
Note though that this is different from imposing the hard role-uniqueness constraint, as the network does this dynamically in a specific context.
%Once the model is quite confident in predicting argument roles, the coupling coefficients $c_{ij}$ in Eq (\ref{eq:intermediate}) 
%trend to zeros on other unexpected roles. 
Still, this is a strong restriction. % in the proposition.
%obtains very limited information from other argument roles in a proposition.

In order to make the model more expressive, we further introduce a global node $\bg^{(t)}$ to incorporate global information about all arguments at the current iterations. The global node is a compressed representation of the entire proposition, and
 used in the prediction of all arguments, thus permitting arbitrary interaction across arguments. 
 %It lets a compressed representation of the entire proposition to be passed across arguments. 
%which can then be used in the structure refinement. 
%argument decisions, so that the coupling coefficients $c_{ij}$ are determined with a balance between local and global information. 
The global node $\bg^{(t)}$ at $t$-th iteration is derived as:
\begin{equation}
    \bg^{(t)} = \bW \bs^{(t)}, \label{eq:global}
\end{equation}
where $\bW \in \cR^{K \times (K \cdot |\cT|)}$. Here, $\bs^{(t)} = \bs_1^{(t)} \oplus \bs_2^{(t)} \oplus \cdots \oplus \bs_{|\cT|}^{(t)} \in \cR^{K \cdot |\cT|}$ is the concatenation of all capsules in the role capsule layer. %Note that the global node gets updated in every iteration.

We append an additional term for the role logits update in Eq (\ref{eq:dynamic}):
\begin{equation}
    b_{j|i}^{(t+1)} = b_{j|i}^{(t)} + \bv_j^{(t)} \bW \bu_{j|i} + \bg^{(t)} \bW_g \bu_{j|i},\label{eq:dynamic_global}
\end{equation}
where $\bW \in \cR^{K \times K}$ and $\bW_g \in \cR^{K \times K}$.

\section{Refinement}\label{sec:refinement}
The dynamic routing process can be seen as iterative role refinement. Concretely, the coupling coefficients $\bc_{i}^{(t)}$ in Eq (\ref{eq:couple}) can be interpreted as the predicted distribution of the semantic roles for word $x_i$ in Eq (\ref{eq:prob}) at $t$-th iteration:
\begin{equation}
    \bc_i^{(t)} = P^{(t)}(\cdot|x, i, p).
\end{equation}
Since dynamic routing is an iterative process, semantic role distribution $\bc_i^{(t)}$ in iteration $t$ will affect the semantic role distribution $\bc_i^{(t+1)}$ in next iteration $t+1$:
\begin{equation}
    \bc_i^{(t+1)} = f (\bc_i^{(t)}, x, i, p),
\end{equation}
where $f(\cdot)$ denotes the refinement function defined by the operations in each dynamic routing iteration.

\section{Training}
We minimize the following loss function $\cL(\theta)$:
\begin{equation}%\small
    \cL(\theta) \!=\! -\frac{1}{n}\!\sum_i^n\log P^{(T)}(y_i|x,i,p; \theta) \!+\! \lambda||\theta||^2_2, \label{eq:single_loss}
\end{equation}
where $\lambda$ is a hyper-parameter for the regularization term and $P^{(T)}(y_i|x,i,p; \theta) =  \bc_i^{(T)}$. 
%The probability distribution $P^{(T)}(\cdot|x,i,p;\theta)$ of the semantic role type for word $x_i$ is the coupling coefficients $\bc_i^&{(T)}$ in $T$-th iteration:
%\begin{align}
%    P(\cdot|x,i,p; \theta) &=  \bc_i^{(T)}\\
%    &= \softmax(\bb_{\cdot|i}^{(T)}),
%\end{align}
%where $\bb_{\cdot|i}^{(T)}$ is the semantic role logits in $T$-th iteration, which is derived via Algorithm \ref{alg:dynamic}.\Ivan{If you add notation $t$ above, you should drop this, and just write $\bc_i^&{(T)}$ } 
Unlike  standard refinement methods \cite{Belanger2017EndtoEndLF,lee-etal-2018-deterministic} which sum losses across all refinement iterations, our loss is only based on the prediction made at the last iteration. This encourages the model to rely on the refinement process rather than do it in one shot.  

Our baseline model is trained analogously, but using the cross-entropy for the independent classifiers (Eq ~(\ref{eq:prob})).
%In such a setting, our model performs weak in the beginning and inclines to obtain the best performance after proceeding the predefined number of iterations.

\paragraph{Uniqueness Constraint Assumption} As we discussed earlier, for a given target predicate, each semantic role  will  typically appear at most once. To encode this intuition, we propose another loss term $\cL_u(\theta)$:
\begin{equation}
    \cL_u(\theta) = \frac{1}{|\cT|}\sum_j \log \softmax(\bb_{j|\cdot}^{(T)}),
\end{equation}
where $\bb_{j|\cdot}^{(T)}$ are the semantic role logits in $T$-th iteration. Thus, the final loss $\cL^*(\theta)$ is the combination of the two losses:
\begin{equation}
    \cL^*(\theta) = \cL(\theta) + \eta\cL_u(\theta),\label{eq:double_loss}
\end{equation}
where $\eta$ is a discount coefficient.

\section{Experiments}
\subsection{Datasets \& Configuration}
we conduct experiments on CoNLL-2009 \cite{conll2009} for all languages, including Catalan (Ca), Chinese (Zh), Czech (Cz), English (En), German (De), Japanese
%\footnote{Much work \cite{ouchi2015joint,ouchi2017neural,matsubayashi2018distance} 
 %on Japanese semantic role labeling task uses Naist Text Corpus \cite{iida2007annotating} as it contains more training instances.} 
 % I think it looks super odd to say that. The same is certainly true for German (the dependency conversion is not very good)
 (Jp) and Spanish (Es). As standard we do not consider predicate identification: the predicate position is provided as an input feature. We use the labeled `semantic' F1 which jointly scores sense prediction and role labeling.

%Labeled F1 measurement on both argument recognition and sense disambiguation is employed for evaluating our model.

We use ELMo \cite{peters-etal-2018-deep} (dimension $d_e$ as 1024) for English, and FastText embeddings~\cite{grave-etal-2018-learning} (dimension $d_e$ as 300) for all other languages on both the baseline model and the proposed CapsuleNet. LSTM state dimension $d_l$ is 500. Capsule size $K$ is 16. Batch size is 32. The coefficient for the regularization term $\lambda$ is $0.0004$. We employ Adam \cite{Kingma2015AdamAM} as the optimizer and the initial learning rate $\alpha$ is set to 0.0001. Syntactic information is not utilized in our model.%\Ivan{If we have space issues, we can move this to the appendix.}

\begin{table}[t]\small%\setlength{\tabcolsep}{3pt}\small %
\centering
\begin{tabular}{c|c|cccc}%>{\columncolor[gray]{.8}}c|}
\toprule
\boldmath$\eta$&\textbf{\# of Iters}	&	\textbf{P}	&	\textbf{R}	&	\textbf{F}&	\textbf{EM}\\
\midrule
\multirow{3}{*}{1.0}&1	&	\textbf{89.31}	&	88.31	&	88.80	&	63.21\\
&2	&	88.34	&	88.92	&	88.63	&	62.80\\
&3	&	88.92	&	\textbf{88.97}	&	\textbf{88.94}	&	\textbf{63.51}\\

\midrule
\multirow{3}{*}{0.1}&1	&	89.16	&	87.99	&	88.57	&	62.94\\
&2	&	88.70	&	88.89	&	88.80	&	63.00\\
&3	&	\textbf{89.75}	&	\textbf{89.40}	&	\textbf{89.57}	&	\textbf{66.01}\\

\midrule
\multirow{3}{*}{0.0}&1	&	89.59	&	\textbf{90.06}	&	89.82	&	66.74\\
&2	&	\textbf{89.88}	&	89.97	&	\textbf{89.92}	&	\textbf{66.93}\\
&3	&	89.25	&	89.61	&	89.43	&	65.98\\
\bottomrule
\end{tabular}
\caption{The performance of our model (w/o global node) with different  discount coefficients $\eta$ of loss $\cL^*(\theta)$ on the English development set. EM denotes the ratio of exact match on propositions.}\label{tab:model_selection}
\end{table}
\subsection{Model Selection} \label{sec:model_select}
Table \ref{tab:model_selection} shows the performance of our model trained with loss $\cL^*(\theta)$ for different values of discount coefficient $\eta$ on the English development set. The model achieves the best performance when $\eta$ equals to 0. 
It implies that adding uniqueness constraint on loss actually hurts the performance.
%by reducing the value of discount coefficient $\eta$. The best model with $\eta$ equals to 0.0 obtains 89.92\% F1, while the best model with $\eta$ equals to 1.0 obtains 88.94\% F1. It implies that adding uniqueness constraint on loss actually hurts the performance. % IT: It was a bit too wordy and we are out of space
Thus, we use the loss $\cL^*(\theta)$ with $\eta$ equals to 0 in the rest of the experiments, which is equivalent to the loss $\cL(\theta)$ in Eq (\ref{eq:single_loss}). We also observe that the model with 2 refinement iterations performs the best (89.92\% F1).

\begin{table}[t]\small%\setlength{\tabcolsep}{3pt}\small %
\centering
\begin{tabular}{l|cc}%>{\columncolor[gray]{.8}}c|}
\toprule
\textbf{Models}&\textbf{Test}&\textbf{Ood}\\
\midrule

%\citet{bjorkelund2010high}	&	 86.90	&75.70&	-\\% (global) 
\citet{lei2015high}	&	 86.60&75.60\\% (local) 
\citet{fitzgerald2015semantic}	&	 87.70&75.50\\% (ensemble) 
\citet{foland2015dependency}	&	 86.00&75.90\\% (global) 
\citet{roth2016neural}	&	 87.90&76.50\\% (ensemble)
\citet{swayamdipta2016greedy}	&	 85.00&-\\% (global) 
\citet{marcheggiani2017simple}	&	 87.70&77.70\\% (syntax-agnostic) 
\citet{marcheggiani2017encoding}	&	89.10&78.90\\% (ensemble)
\citet{he2018syntax}	&	89.50&79.30\\% (syntax-aware) 
\citet{li2018unified}	&	  89.80	&79.80\\% (local) 
\citet{cai2018full}	&	89.60&79.00\\% (local)
\citet{mulcaire2018polyglot} 	&	87.24&-\\
\citet{li2019dependency}$^\dag$	&	90.40&81.50\\
\midrule
Baseline$^\dag$&90.49&81.95\\
\textbf{CapsuleNet$^\dag$ (This Work)}&\textbf{91.06}&\textbf{82.72}\\
% \midrule
% \midrule
% \multicolumn{3}{l}{\textbf{Japanese}}\\
% \midrule
% \citet{bjorkelund-etal-2009-multilingual}	&	\multicolumn{2}{c}{76.30}\\
% \citet{meza-ruiz-riedel-2009-multilingual}	&	\multicolumn{2}{c}{76.00}\\
% \citet{zhao2009multilingual}	&	\multicolumn{2}{c}{78.17}\\
% \citet{zhao-etal-2009-multilingual}	&	\multicolumn{2}{c}{78.15}\\
% \citet{watanabe-etal-2010-structured}	&	\multicolumn{2}{c}{78.69} \\
% \citet{mulcaire2018polyglot} 	&\multicolumn{2}{c}{76.00}\\
% \midrule
% Baseline&\multicolumn{2}{c}{78.50}\\
% \textbf{CapsuleNet (This Work)}&\multicolumn{2}{c}{\textbf{79.81}}\\
\bottomrule
\end{tabular}
\caption{The F1 scores of previous systems on the English test set and out-of-domain (Ood) set. $\dag$ denotes that the model uses Elmo.}\label{tab:overall}
\end{table}

\begin{table*}[t]\small\setlength{\tabcolsep}{2.5pt}\small %
\centering
\begin{tabular}{l|cccc|cccc|cccc}%>{\columncolor[gray]{.8}}c|}
\toprule
\multirow{2}{*}{\textbf{Models}}	&\multicolumn{4}{c|}{\textbf{Development}}&\multicolumn{4}{c|}{\textbf{Test}}&\multicolumn{4}{c}{\textbf{Ood}}\\	
%\cmidrule{2-7}
&\textbf{P}	&	\textbf{R}	&	\textbf{F}&	\textbf{EM}&\textbf{P}	&	\textbf{R}	&	\textbf{F}&	\textbf{EM}&\textbf{P}	&	\textbf{R}	&	\textbf{F}&	\textbf{EM}\\
\midrule
% \midrule
% \multicolumn{7}{l}{\textbf{English Dataset}}\\
% \midrule
\textbf{CapsuleNet SRL} &89.63	&	\textbf{90.25}	&	\textbf{89.94}	&	\textbf{67.15}	&	90.74	&	\textbf{91.38}	&	\textbf{91.06}	&	\textbf{69.90}&\textbf{82.66}	&	\textbf{82.78}	&	\textbf{82.72}	&	\textbf{53.14}\\
\hspace{2mm}w/o Global Node &	89.88	&	89.97	&	89.92	&	66.93	&	\textbf{90.95}	&	91.15	&	91.05	&	69.65&82.45	&	82.27	&	82.36	&	51.95\\
\hspace{2mm}w/o Role Capsule Layer	&	\textbf{89.95}	&	89.64	&	89.80	&	66.70	&	90.82	&	90.66	&	90.74	&	68.46&82.26	&	81.64	&	81.95	&	50.75\\
\hspace{2mm}w/o Word Capsule Layer (Baseline)	&	89.47	&	89.61	&	89.54	&	66.17	&	90.31	&	90.68	&	90.49	&	68.27&82.16	&	81.74	&	81.95	&	50.75\\
% \midrule
% \midrule
% \multicolumn{9}{l}{\textbf{Japanese Dataset}}\\
% \midrule
% \textbf{CapsuleNet SRL}	&	\textbf{86.66}	&	75.04	&	\textbf{80.43}	&	\textbf{42.76}	&	\textbf{86.09}	&	74.38	&	\textbf{79.81}	&	\textbf{41.69}&\multicolumn{4}{c}{-}\\
% \hspace{2mm}w/o Global Node	&	86.28	&	\textbf{75.15}	&	80.33	&	41.98	&	84.92	&	\textbf{74.39}	&	79.31	&	41.27&\multicolumn{4}{c}{-}\\
% \hspace{2mm}w/o Role Capsule Layer	&	84.29	&	75.36	&	79.58	&	40.74	&	83.82	&	73.83	&	78.50	&	38.19&\multicolumn{4}{c}{-}\\
% \hspace{2mm}w/o Word Capsule Layer (Baseline)	&	86.03	&	74.21	&	79.69	&	39.51	&	85.24	&	72.75	&	78.50	&	38.48&\multicolumn{4}{c}{-}\\
\bottomrule
\end{tabular}
\caption{Ablation results in English development and test sets. EM denotes the ratio of the exact match on propositions.}\label{tab:ablation}
\end{table*}

\subsection{Overall Results}
Table \ref{tab:overall} compares our model with previous state-of-the-art SRL systems on English. Some of the systems \cite{lei2015high,li2018unified,cai2018full} only use local features, whereas others \cite{swayamdipta2016greedy} incorporate global information at the expense of greedy decoding. Additionally, a number of systems exploit syntactic information \cite{roth2016neural,marcheggiani2017encoding,he2018syntax}. Some of the  results are obtained with ensemble systems \cite{fitzgerald2015semantic,roth2016neural}.

As we observe, the baseline model (see Section \ref{sec:base}) is quite strong, and outperforms the previous state-of-the-art systems on both in-domain and out-of-domain sets on English. The proposed CapsuleNet outperforms the baseline model on English (e.g. obtaining 91.06\% F1 on the English test set), which shows the effectiveness of the capsule network framework. The improvement on the out-of-domain set implies that our model is robust to domain variations. %CapsuleNet obtains 82.72\% F1 on the English out-of-domain set and achieves a 0.77\% F1 point improvement compared with the baseline system with 81.95\% F1. It implies that our model is robust to domain variations. %, although our model does not incorporate syntax information and does not apply ensemble mechanisms.  \Ivan{But it uses ELMo! Do not over-emphasize the improvement over-non ELMo method. It would have been an automatic reject for me. We cannot emphasize the ELMo diferences (as we do not have the ablation) but we should not pretend it is not the most important factor.} 

\subsection{Ablation}
Table \ref{tab:ablation} gives the performance of models with ablation on some key components, which shows the contribution of each component in our model.
\begin{itemize}
    \item The model \textbf{without the global node} is described in Section \ref{sec:capsule}. 
    \item The model that further \textbf{removes the role capsule layer} takes the mean of capsules $\bu_{j|i}$ in Eq (\ref{eq:word}) of word capsule layer as the semantic role logits $b_{j|i}$:
    \begin{equation}
        b_{j|i} = \frac{1}{K}\sum_k^K \bu_{j|i}^k,\label{eq:ave}
    \end{equation}
    where $K$ denotes the capsule size.
    \item The model that additionally \textbf{removes the word capsule layer} is exactly equivalent to the baseline model described in Section \ref{sec:base}.
\end{itemize}

As we observe, CapsuleNet with all components performs the best on both development and test sets on English. The model without using the global node performs well too. It obtains 91.05\% F1 on the English test set, almost the same performance as full CapsuleNet. But on the English out-of-domain set, without the global node, the performance drops from 82.72\% F1 to 82.36\% F1. It implies that the global node helps in model generalization. Further, once  the role capsule layer is removed, the performance drops sharply.
%(e.g. It obtains 79.58\% F1 score on the Japanese development set, about 0.75\% F1 point less compared to the model without the global node). 
Note that the  model without the role capsule layer does not use refinements and hence does not model argument interaction.
It takes the mean of capsules $\bu_{j|i}$ in the word capsule layer as the semantic role logits $b_{j|i}$ (see Eq (\ref{eq:ave})), and hence could be viewed as an enhanced (`ensembled') version of the baseline model. Note that we only introduced a very limited number of parameters for the dynamic routing mechanism (see Eq (\ref{eq:dynamic}-\ref{eq:dynamic_global})). This suggests that the dynamic routing mechanism does genuinely captures interactions between argument labeling decisions and the performance benefits from the refinement process.
%\Ivan{I'd prefer in text not to mention second digit after "." (The standard style in engineering not to report stat insignificant digits in tex). But that's up to you.}

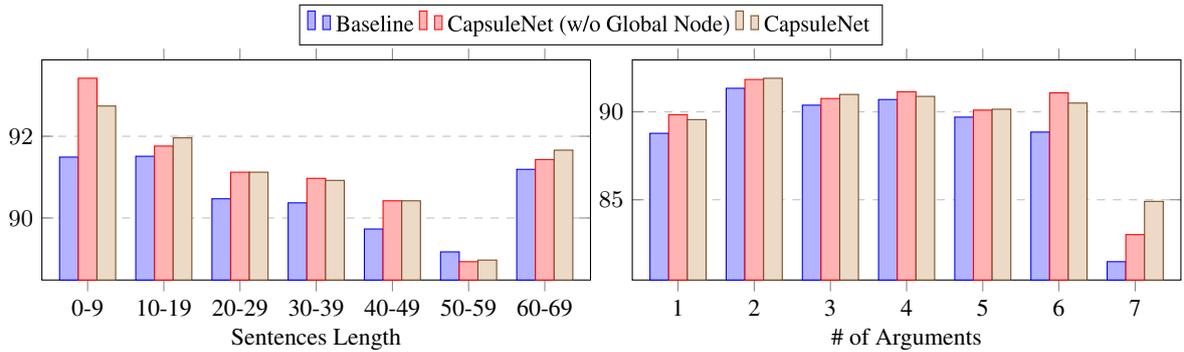
\begin{figure*}[t]\small
  \centering
  %\pgfplotsset{width=0.30\textwidth}
  \includegraphics[width=0.48\textwidth]{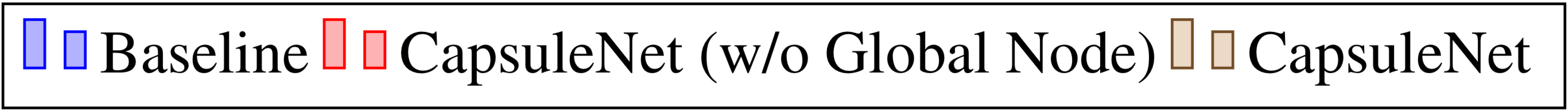}\\
  \vspace*{0mm} 
%  \subfloat[Precision (P)]{\label{fig:error_length}
\begin{tikzpicture}
    \begin{axis}[
    xlabel={Sentences Length},
    %ylabel style={align=center},
    %ylabel=\# of Propositions \\ with Duplicate Arguments,
            ybar,
            ybar=0cm,
            width=0.55\textwidth,
            height=4.5cm,
            legend style={at={(0.5,1.18)},
anchor=north,legend columns=3,font=\footnotesize},
            ymajorgrids=true,
            grid style=dashed,
            xticklabel style={font=\small},
            yticklabel style={font=\small},
            symbolic x coords={0-9,10-19,20-29,30-39,40-49,50-59,60-69},
            xtick=data,
            %nodes near coords,
            bar width=7pt,
            every node near coord/.append style={font=\tiny},
        ]
        \addplot table [x=sen_len, y=f] {sentence_len_plain_10.txt};
        %\addplot table [x=sen_len, y=f] {sentence_len_ensemble_10.txt};
        \addplot table [x=sen_len, y=f] {sentence_len_cap_10.txt};
        \addplot table [x=sen_len, y=f] {sentence_len_cap_with_global_10.txt};
    \end{axis}
\end{tikzpicture}
%}
\hspace{-0.5em}
%  \subfloat[Precision (P)]{\label{fig:error_argnum}
\begin{tikzpicture}
    \begin{axis}[
    xlabel={\# of Arguments},
    %ylabel style={align=center},
    %ylabel=\# of Propositions \\ with Duplicate Arguments,
            ybar,
            ybar=0cm,
            width=0.55\textwidth,
            height=4.5cm,
            legend style={at={(0.5,1.18)},
anchor=north,legend columns=3,font=\footnotesize},
            ymajorgrids=true,
            grid style=dashed,
            xticklabel style={font=\small},
            yticklabel style={font=\small},
            symbolic x coords={1,2,3,4,5,6,7},
            xtick=data,
            %nodes near coords,
            bar width=7pt,
            every node near coord/.append style={font=\tiny},
        ]
        \addplot table [red!20!black,fill=red!80!white,x=arg_num, y=f] {arg_num_plain.txt};
        %\addplot table [x=arg_num, y=f] {arg_num_ensemble.txt};
        \addplot table [x=arg_num, y=f] {arg_num_cap.txt};
        \addplot table [x=arg_num, y=f] {arg_num_cap_with_global.txt};
        %\legend{\testsc{Baseline},  \testsc{CapsuleNet (w/o Global Node)},\testsc{CapsuleNet}}
    \end{axis}
\end{tikzpicture}
%}
\caption{The F1 scores of different models while varying sentence length and argument number on the English test set. }\label{fig:error_analysis}
\end{figure*}

\begin{table}[t]\small
\centering
%    \begin{subtable}
    \begin{tabular}{c|r}
    \toprule
      \textbf{Sent Len} &   \textbf{\# of Props}\\
      \midrule
      0 - 9  & 181\\
      10 - 19 & 1,806\\
      20 - 29 & 3,514\\
      30 - 39 & 3,102\\
      40 - 49 & 1,383\\
      50 - 59 & 391\\
      60 - 69 & 121\\
        \bottomrule
    \end{tabular}
 %   \end{subtable}
  %  \begin{subtable}
    \begin{tabular}{c|r}
    \toprule
      \textbf{\# of Args} &   \textbf{\# of Props}\\
      \midrule
      1  & 2,591\\
      2 & 4,497\\
      3 & 2,189\\
      4 & 889\\
      5 & 259\\
      6 & 39\\
      7 & 7\\
        \bottomrule
    \end{tabular}
 %   \end{subtable}
    \caption{Numbers of propositions with different sentence lengths and different numbers of arguments on the English test set.}
    \label{tab:stat}
\end{table}

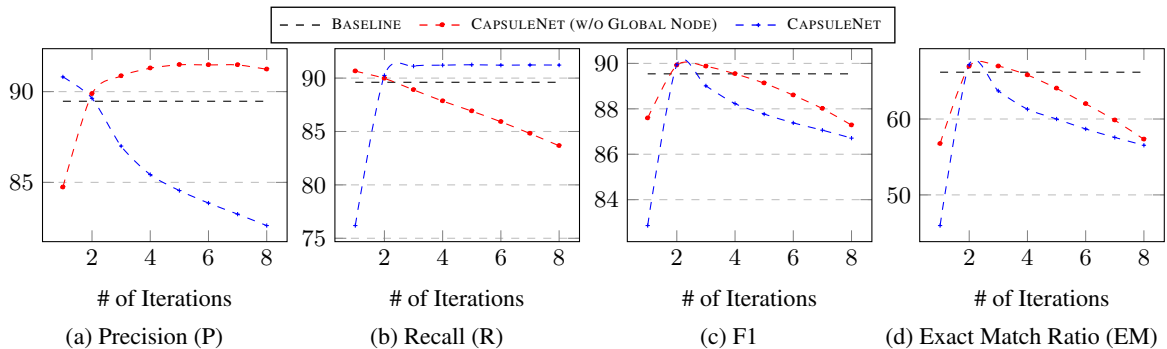
\begin{figure*}[t]\small
  \centering
  \pgfplotsset{width=0.30\textwidth}
  \begin{tikzpicture}
\begin{customlegend}[legend columns=5,legend style={font=\tiny,line width=.5pt,mark size=.8pt,
            /tikz/every even column/.append style={column sep=0.5em}},
        legend entries={\textsc{Baseline} ,
                        %\textsc{Enhanced} ,
                        \textsc{CapsuleNet (w/o Global Node)} ,
                        \textsc{CapsuleNet}
                        }]
        \addlegendimage{black,dashed,mark=None}
        %\addlegendimage{blue,dashed,mark=None}  
        \addlegendimage{red,dashed,mark=otimes*}
        \addlegendimage{blue,dashed,mark=+}
        \end{customlegend}
\end{tikzpicture}\\
\vspace*{-3.3mm} 
\subfloat[Precision (P)]{
  \begin{tikzpicture}
    \begin{axis}[
    xlabel={\# of Iterations},
    legend entries={},
    mark size=0.8pt,
    ymajorgrids=true,
    grid style=dashed,
    legend pos= south east,
    legend style={font=\footnotesize,line width=.5pt,mark size=.2pt,
            /tikz/every even column/.append style={column sep=0.5em}},
            smooth,
    ]
%    \addplot [black,mark=square] table [x index=0, y index=8] {PM_length_analyse.txt};

    \addplot [black,dashed,mark=None] table [x index=0, y index=1] {dev_P.txt};
    %\addplot [blue,dashed,mark=None] table [x index=0, y index=2] {dev_P.txt};
    \addplot [red,dashed,mark=otimes*] table [x index=0, y index=3] {dev_P.txt};
    \addplot [blue,dashed,mark=+] table [x index=0, y index=4] {dev_P.txt};
    \end{axis}
\end{tikzpicture}
}
\hspace{-0.5em}
\subfloat[Recall (R)]{
  \begin{tikzpicture}
    \begin{axis}[
    xlabel={\# of Iterations},
    legend entries={},
    mark size=0.8pt,
    ymajorgrids=true,
    grid style=dashed,
    legend pos= south east,
    legend style={font=\footnotesize,line width=.5pt,mark size=.2pt,
            /tikz/every even column/.append style={column sep=0.5em}},
            smooth,
    ]
%    \addplot [black,mark=square] table [x index=0, y index=8] {PM_length_analyse.txt};

    \addplot [black,dashed,mark=None] table [x index=0, y index=1] {dev_R.txt};
    %\addplot [blue,dashed,mark=None] table [x index=0, y index=2] {dev_R.txt};
    \addplot [red,dashed,mark=otimes*] table [x index=0, y index=3] {dev_R.txt};
    \addplot [blue,dashed,mark=+] table [x index=0, y index=4] {dev_R.txt};
    \end{axis}
\end{tikzpicture}
}
\hspace{-0.5em}
\subfloat[F1]{
  \begin{tikzpicture}
    \begin{axis}[
    xlabel={\# of Iterations},
    legend entries={},
    mark size=0.8pt,
    ymajorgrids=true,
    grid style=dashed,
    legend pos= south east,
    legend style={font=\footnotesize,line width=.5pt,mark size=.2pt,
            /tikz/every even column/.append style={column sep=0.5em}},
            smooth,
    ]
%    \addplot [black,mark=square] table [x index=0, y index=8] {PM_length_analyse.txt};

    \addplot [black,dashed,mark=None] table [x index=0, y index=1] {dev_F.txt};
    %\addplot [blue,dashed,mark=None] table [x index=0, y index=2] {dev_F.txt};
    \addplot [red,dashed,mark=otimes*] table [x index=0, y index=3] {dev_F.txt};
    \addplot [blue,dashed,mark=+] table [x index=0, y index=4] {dev_F.txt};
    \end{axis}
\end{tikzpicture}
}
\hspace{-0.5em}
\subfloat[Exact Match Ratio (EM)]{
  \begin{tikzpicture}
    \begin{axis}[
    xlabel={\# of Iterations},
    legend entries={},
    mark size=0.8pt,
    ymajorgrids=true,
    grid style=dashed,
    legend pos= south east,
    legend style={font=\footnotesize,line width=.5pt,mark size=.2pt,
            /tikz/every even column/.append style={column sep=0.5em}},
            smooth,
    ]
%    \addplot [black,mark=square] table [x index=0, y index=8] {PM_length_analyse.txt};

    \addplot [black,dashed,mark=None] table [x index=0, y index=1] {dev_EM.txt};
    %\addplot [blue,dashed,mark=None] table [x index=0, y index=2] {dev_EM.txt};
    \addplot [red,dashed,mark=otimes*] table [x index=0, y index=3] {dev_EM.txt};
    \addplot [blue,dashed,mark=+] table [x index=0, y index=4] {dev_EM.txt};
    \end{axis}
\end{tikzpicture}
}
\caption{The performance of different models while varying the number of refinement iterations on the English development set. }\label{fig:refine_interations}
\end{figure*}

\begin{figure}[t]\small
  \centering
  \pgfplotsset{width=0.42\textwidth}

  \begin{tikzpicture}
    \begin{axis}[
    xlabel={\# of Iterations},
    ylabel style={align=center},
    ylabel=\# of Propositions \\ with Duplicate Arguments,
    legend entries={\textsc{Gold} ,
                        %\textsc{Enhanced} ,
                        \textsc{Baseline} ,
                        \textsc{CapsuleNet},
                        \textsc{CapsuleNet (w/o Global Node)}
                        },
    mark size=0.8pt,
    ymajorgrids=true,
    grid style=dashed,
    legend pos= north east,
    legend style={font=\tiny,line width=.5pt,mark size=.2pt,
            at={(0.5,1.21)},
            anchor=north,
            legend columns=2,
            /tikz/every even column/.append style={column sep=0.5em}},
            smooth,
    ]
%    \addplot [black,mark=square] table [x index=0, y index=8] {PM_length_analyse.txt};
    \addplot [orange,dashed,mark=None] table [x index=0, y index=1] {unique_constrain.txt};
    \addplot [black,dashed,mark=None] table [x index=0, y index=2] {unique_constrain.txt};
    %\addplot [blue,dashed,mark=None] table [x index=0, y index=3] {unique_constrain.txt};
    \addplot [red,dashed,mark=otimes*] table [x index=0, y index=4] {unique_constrain.txt};
    \addplot [blue,dashed,mark=+] table [x index=0, y index=5] {unique_constrain.txt};
    \end{axis}
\end{tikzpicture}

\caption{The number of propositions with duplicate arguments for various numbers of refinement iterations on the English development set. }\label{fig:duplicate_arguments}
\end{figure}
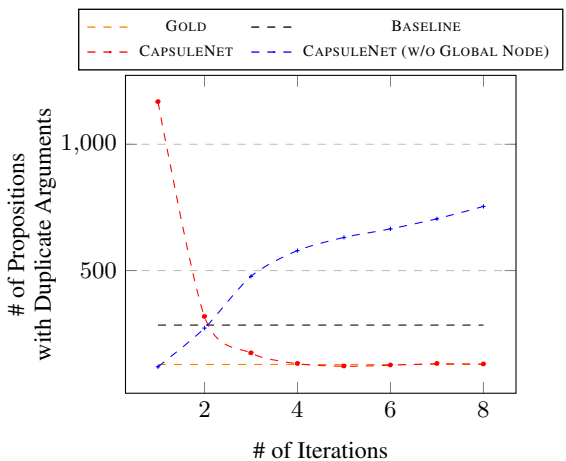

\begin{figure*}[t]\small
    \centering
    \subfloat[Capsule Network w/o Global Node]{
    \includegraphics[width=0.48\textwidth]{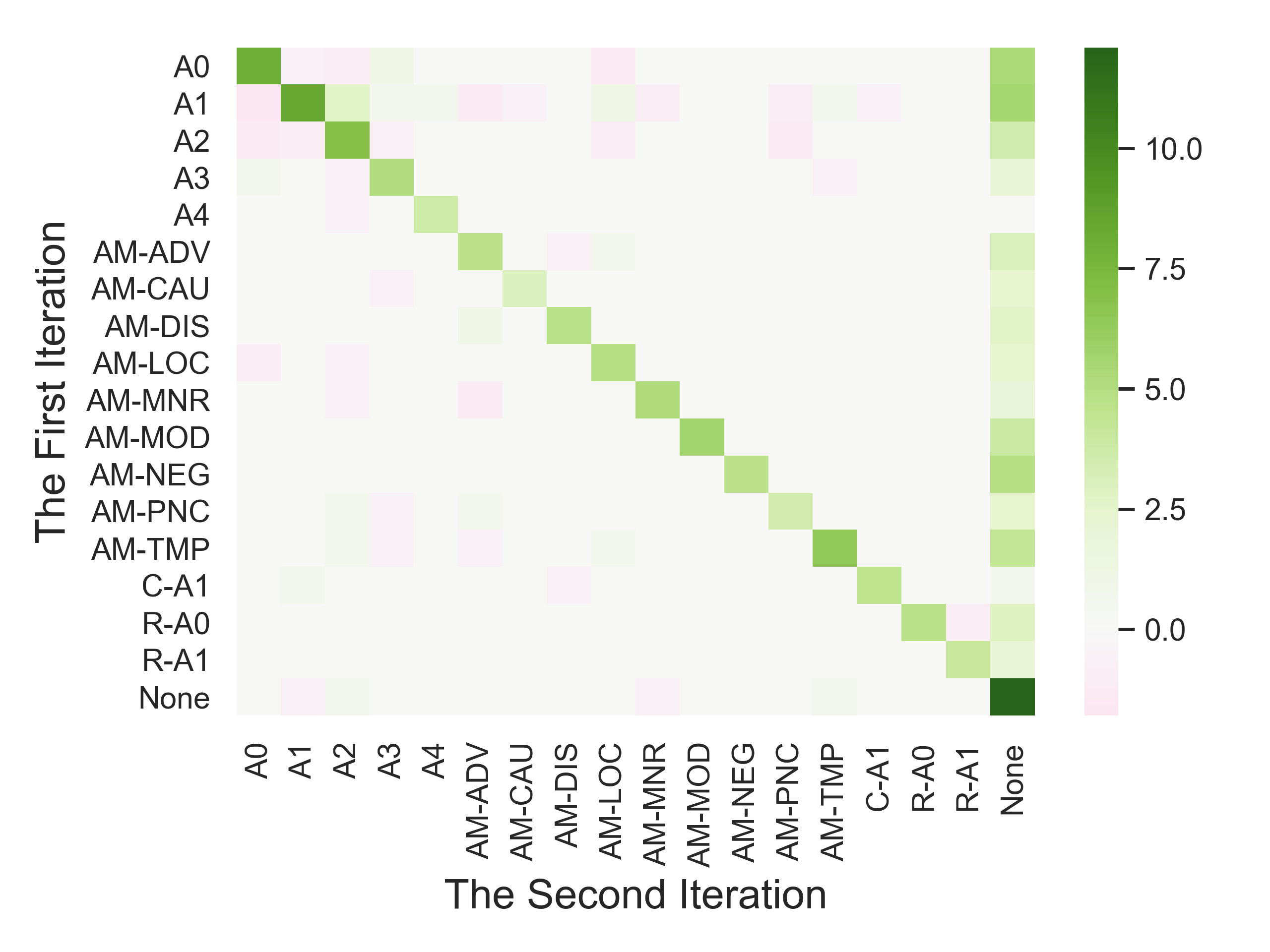}
    }
    \subfloat[Capsule Network with Global Node]{
    \includegraphics[width=0.48\textwidth]{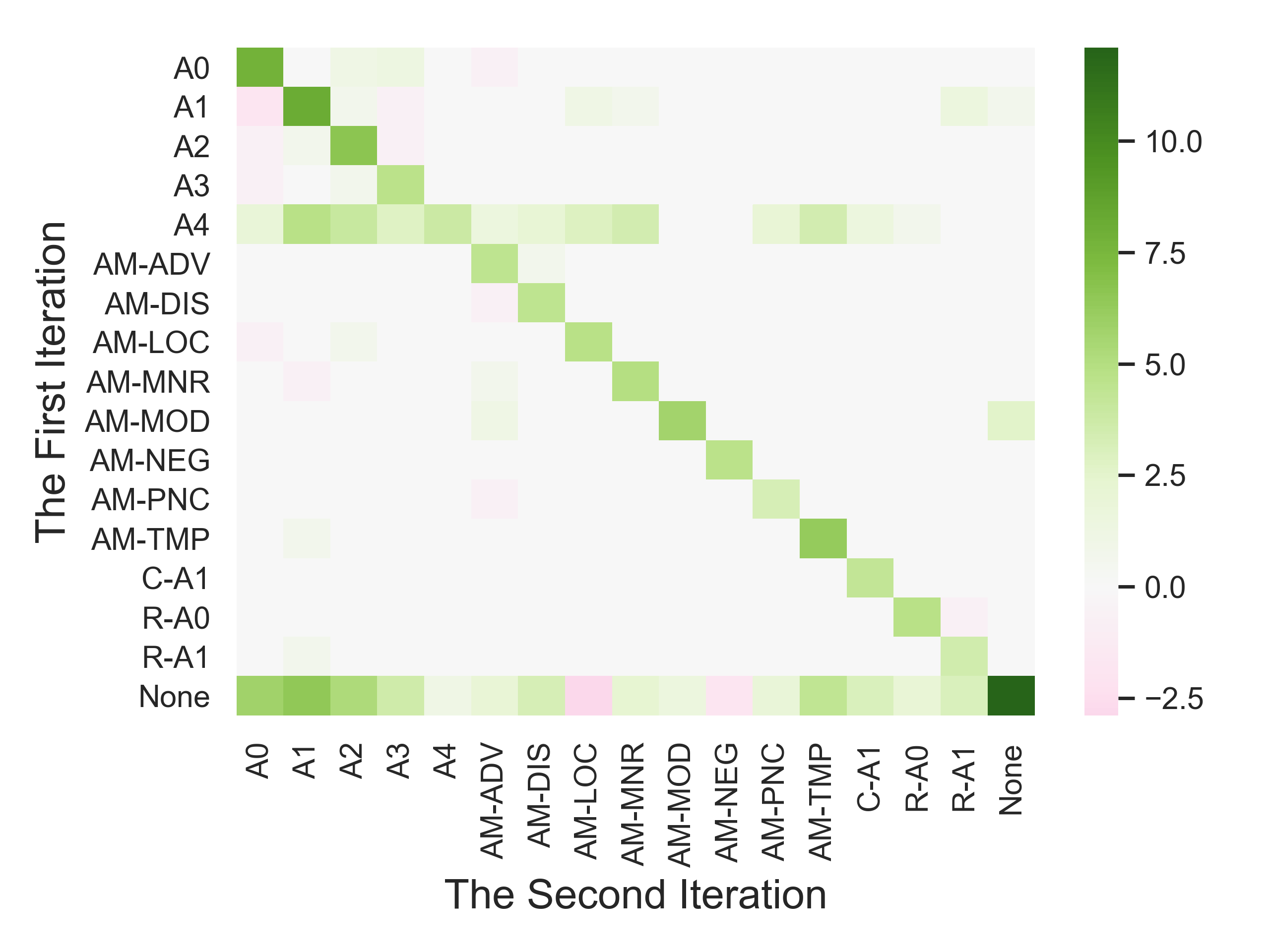}
    }
    \caption{Changes in labeling between consecutive iterations  on the English development set. Only argument types appearing more than 50 times are listed. ``None'' type denotes ``NOT an argument''. Green and red nodes denote the numbers of correct and wrong role transitions have been made respectively. The numbers are transformed into log space.}%The darker the color is, the more the transitions take place.
    \label{fig:transition}
\end{figure*}

\begin{table*}[t]\small
\centering
 \begin{tabular}{l| c c c c c c c | c} 
 \toprule
\textbf{Model} 	&	\textbf{Ja}	&	\textbf{Es}	&	\textbf{Ca}	&	\textbf{De}	&	\textbf{Cz}	&	\textbf{Zh}	&	\textbf{En}	&	\textbf{Avg.}\\
\midrule
Previous Best Single Model	&	78.69	&	80.50	&	80.32	&	\textbf{80.10}	&	86.02	&	\textbf{84.30}	&	90.40	&	82.90\\
Baseline Model	&	80.12	&	81.00	&	81.39	&	76.01	&	87.79	&	81.05	&	90.49	&	82.55\\
\textbf{CapsuleNet SRL (This Work)}	&	\textbf{81.26}	&	\textbf{81.32}	&	\textbf{81.65}	&	76.44	&	\textbf{88.08}	&	81.65	&	\textbf{91.06}	&	\textbf{83.07}\\
 \bottomrule
 %Best Improvement & +0.86 & +0.33 &+1.81 & +0.56  & +0.20 &   +0.99  & +0.17\\
 %\hline
 \end{tabular}
\caption {Labeled F1 score (including senses) for all languages on the CoNLL-2009 in-domain test sets. For previous best result, Japanese (Ja) is from~\newcite{watanabe-etal-2010-structured}, Catalan (Ca) is from~\newcite{zhao2009multilingual}, Spanish (Es) and German (De) are from~\newcite{roth2016neural}, Czech (Cz) is from~\newcite{henderson-etal-2013-multilingual}, Chinese (Zh) is from~\newcite{cai2018full} and English (En) is from~\newcite{li2019dependency}.}
 \label{tab:mul}
\end{table*}

\subsection{Error Analysis}
The performance of our model while varying the sentence length and the number of arguments per proposition is shown in Figure \ref{fig:error_analysis}. The statistics of the subsets are  in Table \ref{tab:stat}. Our model consistently outperforms the baseline model, except on  sentences of between 50 and 59 words. Note that the subset is small (only 391 sentences), so 
the effect may be random.
%As that subset is quite small, consisting of only 391 propositions,
%the outlier may be caused by the data bias. 

\subsection{Effect of Refinement}

\subsubsection{Refinement Performance}
%. Since the dynamic routing mechanism could be viewed as a refinement process,
We take a CapsuleNet trained to perform refinements in two iterations and vary the number of iterations at test time (Figure \ref{fig:refine_interations}). 
%By given a trained CapsuleNet model, we could run any number of iterations of dynamic routing (refinement) process. 
 %illustrates the ne of CapsuleNet. 
 CapsuleNets with and without the global node perform very differently. CapsuleNet without the global node reduces the recall and increases the precision with every iteration. This corresponds to our expectations: the version of the network is only capable of modeling competition and hence continues with this filtering process.
%learns how to filter out wrong arguments.
%In each refinement step, each capsule in word capsule layer interact and compete with other capsules to produce a new role capsule layer, leading to a soft uniqueness constraint on arguments. 
%In such a procedure, more capsules will fall into the role ``None'' (role ``None'' denotes ``NOT an arugment''). Thus, the precision goes up whereas the recall goes down.
In contrast, CapsuleNet with the global node reduces the precision as the number of iteration grows.  This model is less transparent, so it harder to see why it chooses this refinement strategy. As expected, the F1 scores for both models peak at  the second iteration,  the iteration number used in the training phase. The trend for the exact match score is consistent with the F1 score.

%With the help of the global node, in each refinement step, capsules incline to match more global information, leading to less role ``None'' decisions. Thus, the precision goes down whereas the recall goes up. Integrating two counter trends, F1 score peaks at second iteration which is the iteration number we set in our training phase. Exact match ratio is highly consistent with F1 score.

\subsubsection{Duplicate Arguments}
We measure the degree to which the  role uniqueness constraint is violated by both models. We plot the number of violations as the function of the number of iterations (Figure~\ref{fig:duplicate_arguments}).
Recall that the violations do not imply errors in predictions as even the gold-standard data does not always satisfy the constraint (see the orange line in the figure). 
 As expected, CapsuleNet without the global node captures competition and focuses on enforcing this constraint. Interestingly, it converges to keeping the same fraction of violations as the one observed in the gold standard data. 
In contrast, the full model increasingly ignores the constraint in later iterations. This is consistent with the over-generation trend
evident from the precision and recall plots discussed above.

%Some propositions have duplicate arguments. Figure \ref{fig:duplicate_arguments} gives the trend of making prediction with duplicate arguments along with the refinement process of a trained CapsuleNet model. As shown in Table \ref{fig:refine_interations}, CapsuleNet without global node trends to lower recall, whereas CapsuleNet with global node trends to higher recall. Thus, along with refinement process, without global node, capsules making same duplicate roles compete with each other and trend to achieve the role uniqueness constraint. Finally, the number of propositions with duplicate arguments predicted by CapsuleNet (without global node) converges to the number of propositions that the dataset contains. In contrast, CapsuleNet with global node trends to cover more role types, leading to increasing number of propositions with duplicate arguments along with the refinement process.

Figure \ref{fig:transition} illustrates how many  roles get changed between consecutive iterations of CapsuleNet, with and without the global node. Green indicates how many correct changes have been made, while red shows how many errors  have been introduced. Since the number of changes is very large, we represent the non-zero number $q$ in the log space $\tilde{q} = \sign(q) \log(|q|)$. %Here, function $\sign(\cdot)$ returns the value sign. \footnote{Keep as zero, if the number $q$ is zero.} %\xchen{Shoud we use short representation of this equation?}
% \begin{equation}
% \tilde{q}=\left\{
% \begin{aligned}
% &\log(q), &q>0 \\
% &0, &q=0\\
% &-\log(-q), &q<0
% \end{aligned}
% \right.
% \end{equation}

As we expected, for both models, the majority of changes are correct, leading to better overall performance. We can again see the same trends. CapsuleNet without the global node tends to filter our arguments by changing them to  ``None''. The reverse is true for the full model.
%leading to a higher precision and a lower recall. For CapsuleNet with global node, a large proportion of transitions are made to alter role ``None'' to other argument roles, leading to a higher recall and a lower precision.

\subsection{Multilingual Results}
Table~\ref{tab:mul} gives the results of the proposed CapsuleNet SRL (with global node) on the in-domain test sets of all languages from CoNLL-2009. As shown in Table \ref{tab:mul}, the proposed model consistently outperforms the non-refinement baseline model and achieves state-of-the-art performance on Catalan (Ca), Czech (Cz), English (En), Japanese (Jp) and Spanish (Es).  Interestingly, the effectiveness of the refinement method does not seem to be dependent on the dataset size: the improvements on the smallest (Japanese) and the largest datasets (English) are among the largest.

\section{Additional Related Work}

The capsule networks have been recently applied to a number of NLP tasks \cite{xiao-etal-2018-mcapsnet,gong2018information}.
In particular, \newcite{yang-etal-2018-investigating} represent text classification labels by a layer of capsules, and take the capsule actions as the classification probability. Using a similar method, \newcite{xia-etal-2018-zero} perform intent detection with the capsule networks. \newcite{wang2018towards} and \newcite{li2019information} use capsule networks to capture rich features for machine translation.  More closely to our work, \newcite{zhang-etal-2018-attention} and \newcite{Zhang2019MultilabeledRE} adopt the capsule networks for relation extraction. The previous models apply the capsule networks to problems that have a fixed number of components in the output. Their approach cannot be directly applied to our setting.
%It is a computational inefficient to apply their models to our SRL problem, where a role probability is needed for each argument in the sentence. We resolve this by treating the connection logits as role logits.

%There are other SRL models that capture argument interactions. \newcite{watanabe-etal-2010-structured} introduce global factors for their structured perceptron model, and adopt a greedy iterative strategy for inference. Using neural networks, \newcite{Belanger2017EndtoEndLF} assigns a global energy of the proposition that captures argument interactions and perform inference through gradient descent on the output. Both methods improve their corresponding baseline models. 

\section{Conclusions \& Future Work}
State-of-the-art dependency SRL methods do not account for any interaction between role labeling decisions. 
We propose an iterative approach to SRL. In each iteration, we refine both predictions of the semantic structure 
(i.e. a discrete structure) and also a distributed representation of the proposition (i.e. the predicate and its predicted arguments). 
The iterative refinement process lets the model capture interactions between the decisions.
We relied on the capsule networks to operationalize this intuition. We demonstrate that our model is effective, and results in improvements over a strong factorized baseline and state-of-the-art results on standard benchmarks for 5 languages (Catalan, Czech, English, Japanese and Spanish) from CoNLL-2019. In future work, we would like to extend the approach to modeling interaction between multiple predicate-argument structures in a sentence as well as to other semantic formalisms (e.g., abstract meaning representations~\cite{banarescu2013abstract}).

\section*{Acknowledgments}
We thank Diego Marcheggiani, Jonathan Mallinson and Philip Williams for constructive feedback and suggestions, as well as anonymous reviewers for their comments. The project was supported by the European Research Council (ERC StG BroadSem 678254) and the Dutch National Science Foundation (NWO VIDI 639.022.518).

\bibliography{acl2019}
\bibliographystyle{acl_natbib}

\end{document}